\documentclass[conference]{IEEEtran}

\usepackage[utf8]{inputenc}
\usepackage[T1]{fontenc}
\usepackage{graphicx}
\graphicspath{{imagini/}}
\usepackage{amsmath,amssymb}
\usepackage{booktabs}
\usepackage{array}
\usepackage{url}
\usepackage{xcolor}
\usepackage{tabularx}
\usepackage{enumitem}
\usepackage{subcaption}
\usepackage[hidelinks]{hyperref}
\usepackage[american]{babel}
\usepackage{csquotes}
\usepackage[style=apa,backend=biber]{biblatex} 
\DeclareLanguageMapping{american}{american-apa}
\newcolumntype{L}{>{\raggedright\arraybackslash}X}

\begin{document}

\title{VisionAId: An Offline-First Multimodal Android\\
Assistant for People with Visual Impairment,\\
Featuring Personalized Object Retrieval}

\author{\IEEEauthorblockN{Cristian-Gabriel Florea\IEEEauthorrefmark{1} and
Stelian Sp\^{\i}nu\IEEEauthorrefmark{2}}
\IEEEauthorblockA{The Military Technical Academy ``Ferdinand~I'',
Bucharest, Romania\\
\IEEEauthorrefmark{1}cristian.florea@mta.ro \quad
\IEEEauthorrefmark{2}stelian.spinu@mta.ro}
\thanks{A preliminary version of this work was presented at the CERC~2026
International Student Conference, Military Technical Academy, Bucharest, Romania.}%
\thanks{The source code and documentation are publicly available at
\url{https://github.com/floreaGabriel/VisionAId}.}}

\maketitle
\thispagestyle{plain} 
\pagestyle{plain}      

\begin{abstract}
Over 285 million people worldwide live with a visual impairment, for whom
everyday tasks such as avoiding obstacles, locating personal belongings,
recognizing familiar faces, or handling cash remain persistent obstacles to
personal autonomy. Existing assistive applications are typically limited to
recognizing predefined categories, depend heavily on cloud connectivity, or
require dedicated hardware. We present \emph{VisionAId}, an Android application
that turns a commodity smartphone into a real-time visual assistant. The system
integrates six on-device deep-learning models (metric monocular depth
estimation, instance segmentation, visual and facial embeddings, face detection,
and a custom banknote detector) running entirely through ONNX Runtime, with an
optional cloud large language model (Google Gemini Flash) used only for narrative
scene description and automatic object labeling. A distinctive contribution is a
\emph{few-shot} pipeline for personal objects: the user photographs an object
from several angles, and the system later locates that specific instance in the
environment, guiding the user toward it with augmented-reality markers, spatial
audio, and distance-proportional haptics. All feedback is multimodal (Romanian
speech synthesis, voice commands, vibration). On a reference device (Samsung
Galaxy S21 Ultra), INT8 quantization reduces depth latency from
$\sim$1200~ms to $\sim$491~ms, the custom banknote detector reaches an mAP@50 of
0.986, and metric depth is calibrated to below 1~cm of error within 3~m.
\end{abstract}


\section{Introduction}
\IEEEPARstart{F}{or} a person who is blind or has low vision, every movement through a
familiar or unfamiliar space requires a constant cognitive effort: memorizing
the precise position of objects, anticipating obstacles, and relying on tactile
or auditory cues to recognize people and everyday items. According to the
\textcite{who2024vision}, more than 285 million people
worldwide are affected by visual impairment, of whom roughly 39 million are
totally blind. For this population, seemingly trivial tasks (estimating the
distance to an obstacle, finding an object placed somewhere at home, identifying
a banknote at checkout, recognizing a person entering a room) remain daily
challenges with a direct impact on personal autonomy and quality of life.

Modern smartphones have evolved into powerful computational platforms equipped
with high-resolution cameras, gyroscopes, accelerometers, and dedicated
graphics processors. Combined with advances in deep learning, this hardware
allows complex models to run directly on the device, reducing dependence on
network connectivity and preserving user privacy. Inference frameworks such as
the Open Neural Network Exchange (ONNX) Runtime~\parencite{kim2022onnxruntime} were
designed specifically to bring
sizeable models to mobile platforms with latencies acceptable for real-time
processing.

Existing assistive solutions, however, exhibit significant limitations. On the
one hand, general-purpose recognition apps such as Seeing AI or Envision
typically offer \emph{category}-level recognition, without any explicit
mechanism for distinguishing specific instances (e.g., telling \emph{the user's}
wallet from any other wallet). On the other hand, hardware-based solutions such
as Bluetooth tags or positioning beacons entail high acquisition and maintenance
costs, require periodic charging, and present practical constraints in everyday
use. These limitations point to a clear gap: a system that combines general
scene perception with the ability to \emph{learn and recognize the user's own
objects}, while running entirely on an ordinary mobile device.

\emph{VisionAId} addresses these requirements through a concrete, integrated
application. It implements a \emph{few-shot} learning capability: the user
photographs a personal object from multiple angles, and the system can later
recognize it in varied environments, guiding the user toward it through voice
instructions and spatial audio. This deep personalization, combined with
monocular depth estimation and 3D localization through augmented reality (AR),
is the principal point of departure from existing solutions, which are limited
to predefined categories. The system integrates six on-device neural networks,
optimized for mobile execution, complemented by the Gemini
API~\parencite{islam2024gemini} cloud service for optional scene description and
automatic object labeling.

\textbf{Contributions.} Beyond integrating existing models and technologies, this
work makes the following contributions:
\begin{enumerate}[leftmargin=*]
\item \textbf{An offline-first multimodal mobile architecture} that integrates
six visual-assistance functions (proximity estimation, scene description,
personal-object registration and retrieval, color identification, face
recognition, and banknote detection) in a single application, with all critical
functions running locally through ONNX Runtime.
\item \textbf{A few-shot pipeline for personal objects} combining instance
segmentation (YOLO11n-Seg), visual embeddings (MobileCLIP), automatic semantic
labeling, and persistent object profiles, going beyond classic detectors
limited to generic categories.
\item \textbf{Automatic embedding-quality validation with an adaptive
similarity threshold} derived from the distribution of intra-class similarities
and recalibrated conservatively at query time.
\item \textbf{A hybrid visual-identification mechanism} that couples categorical
detection (YOLO) with instance identification (MobileCLIP) through keyword
filtering, a combined score, and a CLIP fallback for objects outside the COCO
vocabulary.
\item \textbf{Temporal stabilization and AR guidance} via an exponential moving
average (EMA) and a five-state machine, coupled with ARCore localization and
progressive auditory feedback for step-by-step guidance.
\item \textbf{A custom detector for Romanian banknotes} trained from scratch on
a personally constructed dataset, with a sequential-confirmation counting
strategy.
\item \textbf{On-device optimization and validation}, including INT8
quantization, batch processing, binary embedding storage, and camera sharing
across modules, profiled on a reference device.
\end{enumerate}

To support reproducibility, the complete source code and documentation are
publicly available at \url{https://github.com/floreaGabriel/VisionAId}.

\section{Related Work}
The market for visual-assistance applications has grown substantially in recent
years, alongside advances in computer vision and on-device processing. We review
representative commercial and academic solutions and position our system with
respect to them.

\subsection{Commercial applications}
\textbf{Microsoft Seeing AI} organizes multiple functions into ``channels''
(short-text reading, scene description, barcode product identification, person
recognition, color and light detection, and banknote recognition for several
currencies). Advanced functions rely on Microsoft Cognitive Services in the
cloud. It is iOS-only, offers no metric depth or spatial navigation, does not
allow registering and searching personal objects, and supports neither Romanian
nor the RON currency. \textbf{Be My Eyes} integrates ``Be My AI'', a
GPT-4o-based~\parencite{openai2024gpt4o} assistant answering visual questions from
photographs; it is fully cloud-dependent, processes only static images, and
provides no real-time detection, depth estimation, or AR. \textbf{Google
Lookout} (Android) provides text reading, food-label scanning, scene
description, and limited currency identification using on-device TFLite and
cloud Gemini models, but offers no face recognition, personal-object
registration, AR search, depth estimation, or Romanian support. \textbf{Envision
AI} offers OCR, scene description, face recognition, and a GPT-4o conversational
mode, with advanced features behind a subscription and optional smart glasses.
\textbf{Cash Reader} is dedicated exclusively to banknote recognition across
100+ currencies (including RON), with fully local inference, but covers no other
assistive function. \textbf{Aira} provides professional assistance through video
calls with trained human agents, at a high recurring cost and with full
dependence on connectivity and agent availability.

\subsection{Academic work}
\textcite{okolo2024assistive} present a systematic survey of navigation
aids for people with visual impairment, identifying dependence on dedicated hardware,
high cost, and lack of personalization as recurring limitations; our work adopts
a fully offline-first architecture and adds the personalization their taxonomy
lacks. \textcite{anjom2025objectalert} propose an embedded real-time
obstacle-alert system based on monocular depth with proximity-proportional
audio warnings; in contrast, VisionAId uses \emph{metric} (centimeter-calibrated)
depth within a multifunctional system. \textcite{zhang2019arcore} build
an ARCore-based assistive navigation system using spatial anchors; we reuse the
same 3D-localization technology but apply it to retrieving previously registered
personal objects rather than point-to-point navigation. \textcite{das2025pathfinder} present \emph{PathFinder}, a fully offline mobile
system that finds the longest obstacle-free route from monocular depth; we share
the offline, depth-based approach but employ metric depth within a broader
platform. \textcite{hayath2023voice} develop a voice-interactive
Android indoor-navigation app with ARCore Depth API obstacle detection (the
closest to our technology stack), which we extend beyond navigation toward
personal-object retrieval, face recognition, and banknote detection.

\subsection{Positioning}
Table~\ref{tab:comparison} summarizes the functional comparison. To the best of
our knowledge, none of the studied commercial solutions simultaneously offers
metric depth estimation, personalized AR object search, face recognition, and
Romanian-banknote detection within a single offline-first application. The
personalized retrieval of \emph{the user's own} object with spatial AR guidance
is the least-covered capability among existing solutions, and is the central
differentiator of our system.

\begin{table}[t]
\centering
\caption{Functional comparison of existing solutions and VisionAId.}
\label{tab:comparison}
\footnotesize
\setlength{\tabcolsep}{3pt}
\begin{tabular}{@{}lccccccc@{}}
\toprule
\textbf{Feature} & \rotatebox{60}{Seeing AI} & \rotatebox{60}{Be My Eyes} &
\rotatebox{60}{Lookout} & \rotatebox{60}{Envision} & \rotatebox{60}{Cash Reader}
& \rotatebox{60}{Aira} & \rotatebox{60}{\textbf{VisionAId}} \\
\midrule
Real-time detection      & $\sim$ & --     & $\sim$ & --     & --     & --     & \checkmark \\
Metric depth             & --     & --     & --     & --     & --     & --     & \checkmark \\
Personal-object reg.     & --     & --     & --     & --     & --     & --     & \checkmark \\
AR search + guidance     & --     & --     & --     & --     & --     & --     & \checkmark \\
Face recognition         & \checkmark & -- & --     & \checkmark & -- & --     & \checkmark \\
RON banknotes            & --     & --     & $\sim$ & --     & \checkmark & --   & \checkmark \\
Offline operation        & $\sim$ & --     & $\sim$ & $\sim$ & \checkmark & --   & \checkmark \\
Romanian support         & --     & --     & --     & --     & --     & --     & \checkmark \\
\bottomrule
\end{tabular}
\end{table}

\section{Methods}

\subsection{System architecture}
Unlike traditional client--server applications, VisionAId is designed as a
\emph{standalone} mobile application in which the entire processing pipeline
(video capture, ML inference, data persistence, and feedback generation) runs
locally on the user's device. The only external component is the Google Gemini
Flash cloud API, used optionally for scene description and object labeling; Romanian
voice recognition may also fall back to the Android cloud
speech recognizer when the on-device variant is unsupported. This choice
follows a fundamental accessibility requirement: a user who is blind must be able to
use the application at any time, including without connectivity. Consequently,
all critical functions (obstacle detection, object search, face recognition,
banknote detection) work fully offline.

\begin{figure}[t]
\centering
\includegraphics[width=\columnwidth]{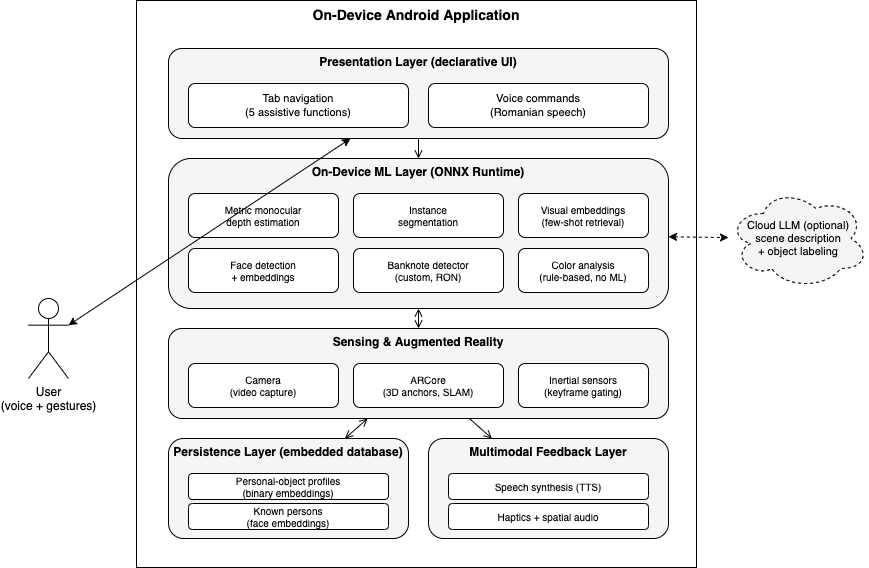}
\caption{Overall layered architecture of VisionAId. All perception runs
on-device; the cloud language model is an optional component used only for
narrative scene description and object labeling.}
\label{fig:architecture}
\end{figure}

The application is implemented natively in Kotlin~2.0 (minimum API~24,
target~API~36) with a Jetpack Compose declarative UI. It is organized into four
software layers: a \emph{presentation} layer (hosting tab
navigation and a voice-command manager), an \emph{ML processing} layer
(six on-device models, each wrapped behind a uniform
initialize/process/release interface), a \emph{persistence} layer
(an embedded relational database storing embeddings as binary blobs), and a
\emph{feedback} layer (speech synthesis, haptics, and sound). Fig.~\ref{fig:architecture} shows
the overall architecture.

Two design principles are central. First, \emph{lazy loading}: models are not
loaded at startup but only on first use of the associated function, keeping
startup under 3~s and limiting memory to the models needed by the active tab; on
screen closure, models are released after a 500~ms delay (\emph{deferred
closing}) to avoid use-after-free errors from inference tasks still running on
background threads. Second, an \emph{always-alive camera} pattern: the CameraX
instance remains in the Compose tree even when the camera tab is hidden (with
$\alpha=0$), eliminating a 1--2~s reinitialization latency on tab switches, while
ML processing is suspended via a screen-visibility flag. The camera is
fully removed from the composition only when sub-screens with their own camera
session are active (object registration, AR search, face registration), because
Android allows a single active camera session at a time.

Embeddings are stored in Room as binary blobs rather than serialized JSON:
deserializing a 512-float vector from JSON took $\sim$150~ms, whereas reading a
binary blob takes under 1~ms, a $>$150$\times$ improvement that is critical for
the frequent comparisons performed during real-time search. The on-device models
total $\sim$210~MB (Table~\ref{tab:models}).

\begin{table}[t]
\centering
\caption{On-device ML models integrated in VisionAId.}
\label{tab:models}
\footnotesize
\begin{tabular}{@{}lllc@{}}
\toprule
\textbf{Model} & \textbf{Backend} & \textbf{Use} & \textbf{Size} \\
\midrule
Depth Anything V2 (INT8) & ONNX & Proximity (Tab 0) & 35~MB \\
YOLO11n-Seg              & ONNX & Segmentation + detect & 12~MB \\
MobileCLIP2-S2           & ONNX & Visual embeddings & 143~MB \\
YuNet                    & ONNX & Face detection & 0.23~MB \\
MobileFaceNet            & ONNX & Face embeddings & 14~MB \\
YOLO26n (Money)          & ONNX & Banknotes (custom) & 9.8~MB \\
\bottomrule
\end{tabular}
\end{table}

\subsection{Proximity estimation and scene description}
\begin{figure}[t]
\centering
\includegraphics[width=0.55\columnwidth]{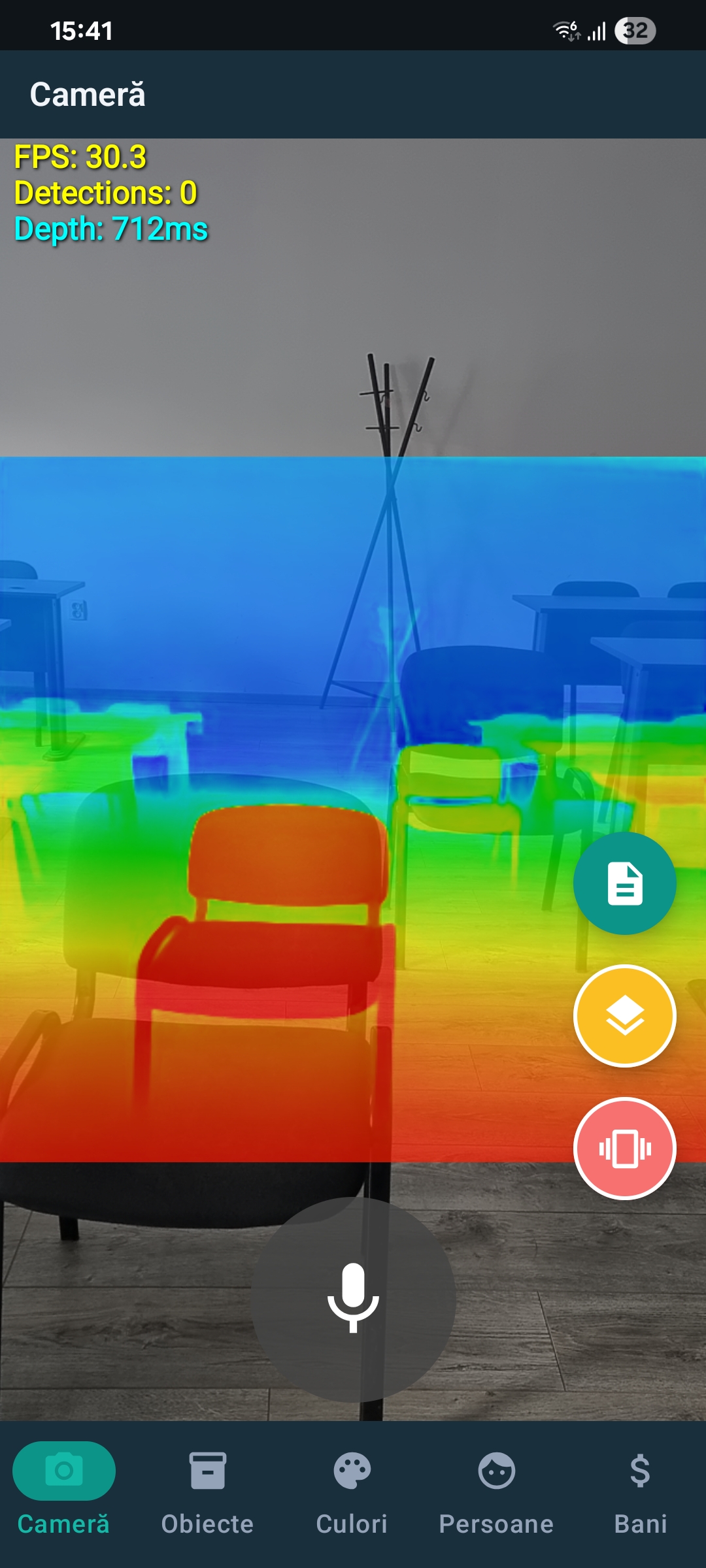}
\caption{Camera module (Tab~0): real-time proximity estimation from the metric
depth map, with warm colors indicating nearby obstacles in the central walking
band.}
\label{fig:camera-screen}
\end{figure}

The camera module (Tab~0) serves a dual purpose: real-time obstacle alerting and
on-demand verbal scene description. An early version ran both a YOLO11n detector
and the depth model on every frame, dropping below 1 frame per second (FPS) on
mid-range devices;
the final design removes YOLO from this tab, keeping only depth estimation for
proximity alerts and delegating scene description to the cloud, which yields a
stable 7--15~FPS. Fig.~\ref{fig:camera-screen} shows the camera module in use.

The depth component wraps the INT8-quantized Depth Anything~V2 Metric Small
model~\parencite{depth_anything_v2}, which takes a $384\times384$ RGB image and
produces a $378\times378$
metric depth map. Preprocessing follows the standard ImageNet protocol (NCHW,
mean/standard-deviation normalization). Inference runs on CPU with four threads
and full graph optimization; NNAPI was deliberately avoided because graph
partitioning of unsupported \texttt{MatMul}/\texttt{Gemm} operators back to CPU
raised latency above 4200~ms on Exynos chipsets.

For obstacle detection it suffices to examine a narrow vertical band of the
frame (25\% width $\times$ 100\% height) aligned with the walking direction, from
which the minimum metric distance is extracted. Keeping the full frame height is
ergonomically motivated: with the phone held at chest level, low obstacles
(stools, boxes, steps) appear only in the lower part of the frame and tall
obstacles in the upper part. The raw model output is scaled by an empirical
calibration factor $f_c=0.55$,
\begin{equation}
d_{\text{cal}} = d_{\text{raw}} \times f_c ,
\end{equation}
which proved sufficiently accurate for proximity alerting (Section~IV-E). A
three-level alert system (\textsc{Danger} $<$0.5~m, \textsc{Warning} 0.5--1.0~m,
\textsc{Caution} 1.0--1.5~m) drives vibration and sound with intensity
proportional to proximity, each level debounced (300/500/800~ms) to prevent
sensory overload; a higher-severity transition fires immediately.

On demand, the user can request a narrative description: the last captured frame
is JPEG-compressed (quality~85, max 1024~px), Base64-encoded, and sent to
Gemini Flash with a Romanian prompt instructing a concise spatial
description, returned through text-to-speech (TTS) in 1--3~s. A local
\emph{sustained-proximity} guidance path also exists: when the central distance
stays below a critical threshold for $\geq$1500~ms (with a 10~s global
cooldown), a frame is sent automatically to obtain a short verbal instruction
about the obstacle and how to avoid it. The cloud model was preferred over local
captioning because on-device captioners produced generic 5--7-word English
descriptions, insufficient for spatial orientation.

\subsection{Personal-object registration}
A frequent difficulty for people with visual impairment is locating \emph{personal}
objects (keys, remote, wallet, medication), which vary enormously between users
and cannot be identified by a generic pretrained detector (e.g., YOLO on COCO
classes~\parencite{lin2014coco}). The registration module solves this by building a
unique \emph{visual profile} for each object: the user films it from several
angles and distances, and the app automatically constructs a set of numerical
embeddings encoding its appearance.

Registration proceeds in three successive phases at different distances
($\sim$30~cm, $\sim$1~m, 2--3~m), targeting $\sim$50 embeddings to capture both
fine texture and overall silhouette. Not every camera frame is useful, so
\emph{keyframes} are selected using inertial sensors: gyroscope angular velocity
is integrated over time, and a frame is captured when the weighted rotation
difference (azimuth dominant) exceeds $15^\circ$, with a 1.5~s timeout forcing
progress. Each candidate is additionally checked for sharpness via the variance
of the discrete Laplacian; frames below threshold are rejected to avoid motion
blur.

Each keyframe is segmented by YOLO11n-Seg~\parencite{khanam2024yolov11}, which removes
the background
(transparent alpha) so that embeddings encode the object rather than its
context; the detected YOLO labels are stored as search \emph{keywords}. If no
object is detected, the system gracefully falls back to the original image. The
segmented image is passed to MobileCLIP2-S2~\parencite{mobileclip2024}, a mobile
variant of the Contrastive Language--Image Pre-training (CLIP)~\parencite{radford2021clip}
family, producing a 512-dimensional
embedding that is L2-normalized for cosine comparison. The clearest captured
frame is sent to Gemini for automatic semantic labeling, returning a structured
response with a Romanian label, a short description, and English
keywords matching COCO classes.

Before storage, the embedding set is validated automatically: the system
computes the mean intra-class cosine
similarity $\bar{s}$ (ideal $0.60\!\leq\!\bar{s}\!\leq\!0.95$), a simulated
retrieval score, and an optimal acceptance threshold
\begin{equation}
\tau^{*} = \max\!\left(P_{10}-0.05,\; \bar{s}-2\sigma\right),\quad
\tau^{*}\in[0.55, 0.75],
\end{equation}
where $P_{10}$ is the 10th percentile of the similarity distribution. The
object-specific $\tau^{*}$ is stored and used adaptively at query time, avoiding
a fixed universal threshold. The validated object is saved as a
personal-object record, with embeddings stored as a binary blob
(little-endian 32-bit floats, preceded by a 32-bit count); for 50
embeddings this occupies $\sim$100~KB versus $\sim$250~KB for JSON, and parses in
under 1~ms.

\subsection{Augmented-reality search}
\begin{figure}[t]
\centering
\begin{subfigure}[b]{0.46\columnwidth}
\centering
\includegraphics[width=\textwidth]{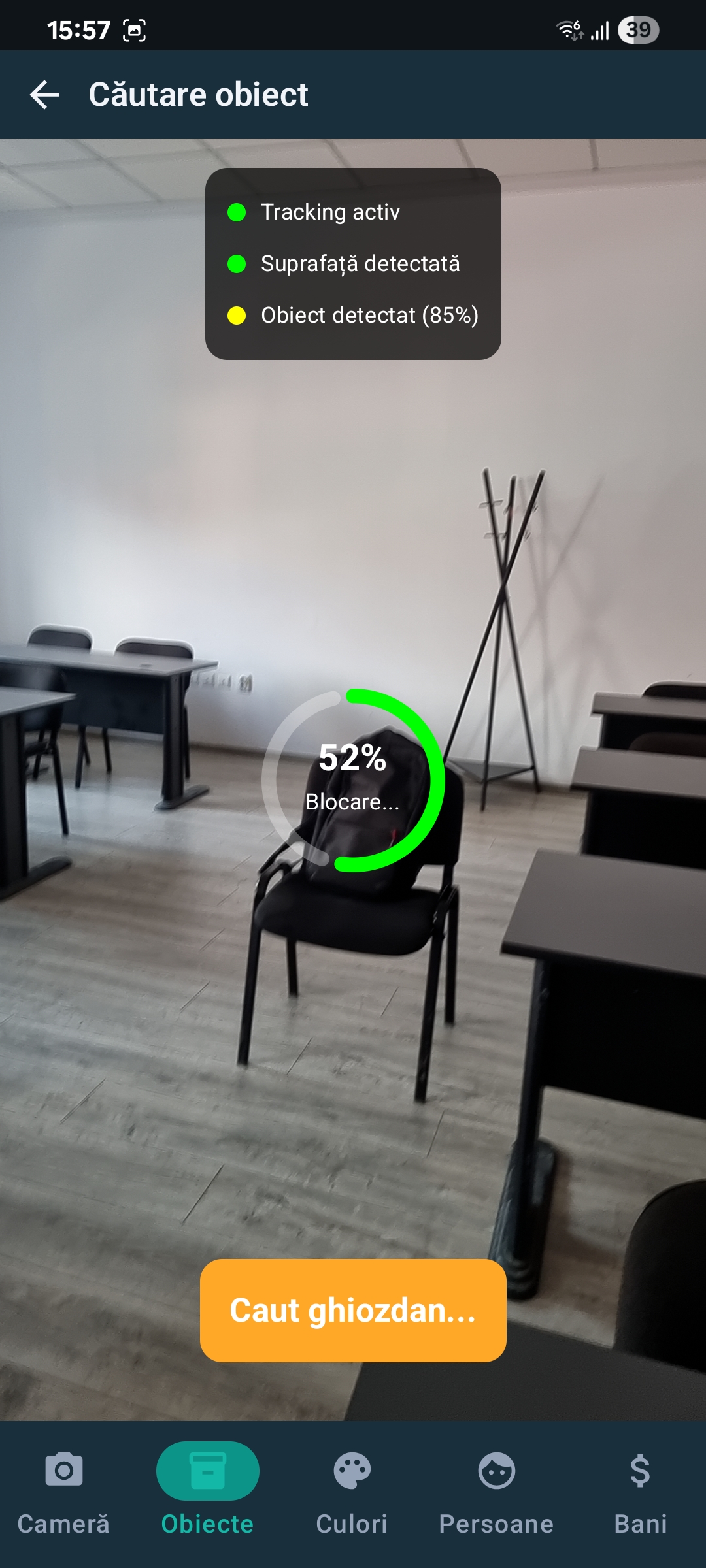}
\caption{Scanning}
\label{fig:ar-scan}
\end{subfigure}
\hfill
\begin{subfigure}[b]{0.46\columnwidth}
\centering
\includegraphics[width=\textwidth]{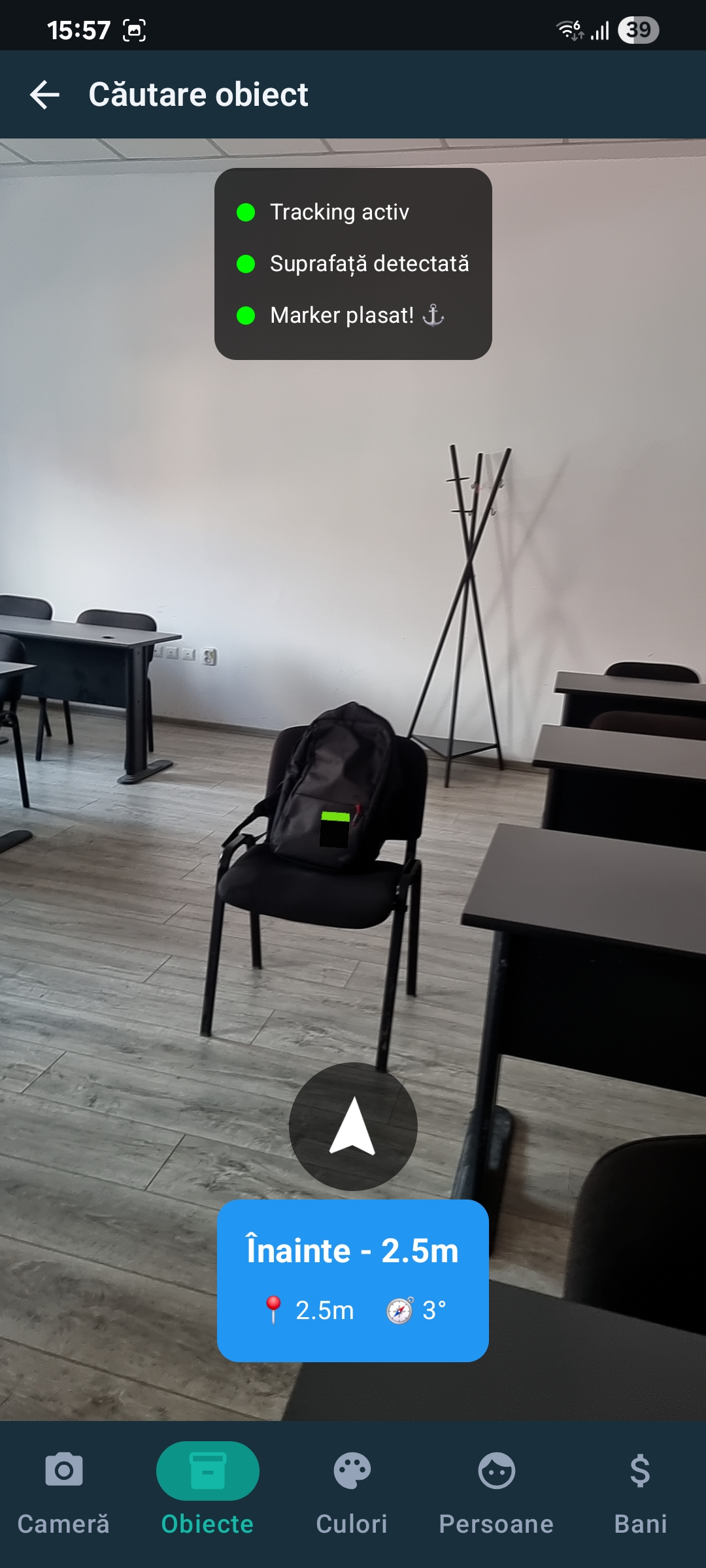}
\caption{Detection and AR marker}
\label{fig:ar-detect}
\end{subfigure}
\caption{The two phases of augmented-reality search: (a)~the system scans the
environment looking for the registered personal object; (b)~once identified, a 3D
marker is anchored on the object and distance-proportional directional guidance
begins.}
\label{fig:ar-search}
\end{figure}

The AR search module is the most complex component, integrating detection and
segmentation (YOLO11n-Seg), visual identification (MobileCLIP), spatial
localization (ARCore), and multimodal guidance. It runs a continuous loop at
$\sim$5~FPS (200~ms minimum interval), with each frame passing through four
stages: (i)~YOLO detection and instance segmentation; (ii)~candidate filtering by
size (area $\in[0.2\%, 50\%]$ of the frame) and category (excluding furniture and
persons), keeping at most two candidates prioritized by keyword match;
(iii)~batched CLIP extraction on the candidate crops; and (iv)~cosine comparison
against the stored profile, feeding an EMA tracker. The per-frame budget is
$\sim$76~ms on the reference device. Fig.~\ref{fig:ar-search} illustrates the two
phases of the search, from environment scanning to marker placement and guidance.

At initialization, the object's embeddings are loaded (binary blob, $<$1~ms), an
L2-normalized centroid is computed, and outliers below the class mean are removed
before recomputing the centroid. For each candidate, the score combines the
centroid similarity with the average over the three most similar profile
embeddings,
\begin{equation}
s_{\text{raw}} = 0.5\,\cos(\hat{e}_{c},\hat{c}) +
0.5\cdot\frac{1}{3}\sum_{k=1}^{3}\cos(\hat{e}_{c},\hat{e}_{(k)}) ,
\end{equation}
and is adjusted by category agreement: a matching YOLO keyword grants a 10\%
bonus, while a mismatch raises the effective acceptance threshold by 0.10. This
hybrid scheme (YOLO narrowing the search space, CLIP discriminating between
instances of the same category) is markedly more robust than either method
alone; a direct center-crop CLIP fallback handles objects outside the COCO
vocabulary.

Single-frame detections are unreliable, so confidence is stabilized through an
exponential-moving-average tracker updated per frame:
\begin{equation}
C_t =
\begin{cases}
\alpha\, s^{2} + (1-\alpha)\,C_{t-1}, & \text{hit } (s \geq \tau^{*}),\\[2pt]
0.95\,C_{t-1}, & \text{skip (no candidate)},\\[2pt]
0.75\,C_{t-1}, & \text{miss},
\end{cases}
\end{equation}
with $\alpha=0.4$. Squaring the score emphasizes high-confidence matches, and the
slow skip decay reflects that a filtered-out person does not mean the target has
disappeared (it may be momentarily occluded). The EMA confidence drives a
five-state machine (\textsc{Scanning}, \textsc{Found}, \textsc{Locked},
\textsc{Lost}, \textsc{Arrived}); in \textsc{Found}, Geiger-counter-style beeps
shorten from 800~ms to 150~ms as confidence rises, intuitively conveying
proximity without visual attention.

When confidence reaches 0.55, a 3D marker is placed via ARCore~\parencite{du2020depthlab}
using a three-level fallback (hit test at the detection, then at screen center, then
instant placement at an estimated 1.5~m). After placement, plane
detection is disabled to save 25--30\% GPU. The module then enters guidance mode:
the camera-to-anchor vector is transformed into camera-local coordinates via the
inverse camera quaternion, from which a guidance angle and Euclidean distance are
computed and converted into Romanian voice instructions every 4~s (e.g.,
``slightly right'', ``ahead, 2.3~m''). Distance-proportional haptics complete the
guidance when the marker is in front and centered. A direct YUV-to-RGB
conversion (ITU-R BT.601), instead of an intermediate JPEG
round-trip, saves 15--25~ms per frame.

\subsection{Color identification}
The color module (Tab~2) deliberately uses no neural network: at 1~Hz it samples
a central $30\times30$-pixel region, converts each pixel to RGB (ITU-R BT.601),
averages the 900 pixels to suppress sensor noise, converts to the
hue--saturation--value (HSV) space, and applies
deterministic rules. Achromatic pixels are classified first,
\begin{equation}
\text{color} =
\begin{cases}
\text{black}, & V < 15\%,\\[2pt]
\text{white}, & S < 15\%,\ V > 80\%,\\[2pt]
\text{gray},  & S < 15\%,
\end{cases}
\end{equation}
followed by chromatic classification over eight hue
intervals, with a light/dark suffix derived from $V$. This yields sub-millisecond
latency, negligible energy use, full predictability, and no model dependency, a
sound trade-off for the eleven base color categories that cover practical needs.

\subsection{Face recognition}
The people module (Tab~3) provides guided registration and real-time
recognition through a two-stage pipeline. \textbf{YuNet}~\parencite{yu2022yunet}
(0.23~MB) detects faces at three strides over 8{,}400 anchors and returns five
facial landmarks, applying non-maximum suppression (NMS) at an
intersection-over-union (IoU) of 0.3 and a 0.5 confidence threshold.
\textbf{MobileFaceNet}~\parencite{chen2018mobilefacenet} (14~MB), chosen over heavier
alternatives such as ArcFace-ViT ($>$100~MB) and FaceNet-Inception ($\sim$90~MB),
then extracts an
L2-normalized 512-dimensional embedding from a $112\times112$ aligned crop. Live
recognition follows a \emph{detect--track--embed--match} loop: faces are tracked
across frames by IoU ($>$0.4), embeddings are re-extracted only every fifth frame
or for new faces, and matches are accepted above a cosine threshold $\tau=0.45$.

Robustness depends on angular diversity, so registration systematically collects
embeddings from eight distinct head angles. The horizontal head rotation (yaw) is
estimated from the landmarks as the nose displacement relative to the eye center,
normalized by inter-ocular distance, and discretized into eight $22.5^\circ$
buckets; an embedding is captured when its bucket is still uncovered. Scanning
stops at 15 embeddings or a 20~s timeout, and the embeddings are stored as binary
blobs per known person.

\subsection{Banknote detection}
Romanian banknotes have similar dimensions and lack distinctive tactile cues
between denominations. The money module (Tab~4) counts banknotes one at a time
using a YOLO26n model~\parencite{chakrabarty2026yolo26} (9.8~MB) trained
\emph{entirely by the author} on a personally built dataset, since no public
dataset of current RON banknotes exists. The model takes a $640\times640$ image,
outputs detections for the eight circulating denominations
(1/5/10/20/50/100/200/500~RON), and applies NMS (IoU~0.45, confidence~0.35).

The dataset was built on Roboflow: banknotes were photographed under varied
lighting, backgrounds, angles, and partial occlusion; manually annotated with
bounding boxes; quality-checked; and iterated over three versions. An important
methodological detail was \emph{iterative class rebalancing}: in v1 the random
split placed the entire 500~RON class in the training set, making it
unevaluable; v3 manually moved 500~RON images to obtain an approximate 70/15/15
split across all three subsets (for the 500~RON class, 76~training / 16~validation
/ 16~test). Augmentation included horizontal flip, $\pm
15^\circ$ rotation, and brightness/saturation/exposure jitter, but \emph{no}
vertical flip (banknotes are not seen upside down).

Training used transfer learning from COCO weights (Ultralytics) on a Tesla~T4:
AdamW ($lr_0=0.001$, weight decay~0.01), cosine annealing, batch~16, image
size~$640$, mixed precision, 120 epochs with early stopping (patience~25), and
close-mosaic over the last 10 epochs. At inference, a \emph{one-at-a-time}
state machine requires three consecutive frames with the same label to confirm a
banknote (and three empty frames to confirm withdrawal), preventing spurious or
double counting; the running total and its denomination breakdown are announced
through TTS.

\subsection{Multimodal feedback}
All modules share a multimodal feedback system. \textbf{Speech synthesis} is
configured for Romanian at $1.15\times$ rate and $0.8\times$ pitch (a slightly
above-normal rate matching screen-reader user preferences~\parencite{asakawa2003} and
a lower pitch improving distinction from ambient voices), and each utterance
flushes the queue to prioritize the most recent information. \textbf{Voice
commands} use a hold-to-talk recognizer with automatic on-device-to-cloud
fallback; recognized text is pattern-matched into navigation,
description, search, save, and control commands, with object names extracted from
parameterized commands. \textbf{Haptics} and low-latency sound effects (flagged
with an accessibility sonification usage) provide
proximity alerts with distance-proportional intensity and variable playback rate.

\section{Results}
All latency measurements were taken on a single reference device, the Samsung
Galaxy S21 Ultra 5G (Exynos~2100, 12~GB RAM, Android~14), representative of the
mid-to-high-end segment, with no dedicated ML accelerator. Each value is the
mean over 50 consecutive frames.


\subsection{Model selection}
For depth estimation, a relative-depth TFLite model was replaced by a metric
ONNX model: although larger ($\sim$100~MB FP32 vs.\ $\sim$50~MB) and slower
($\sim$1200~ms vs.\ $\sim$800~ms on CPU), the metric output eliminates an
error-prone inverse conversion, reducing the error at 1.5~m from $\pm$30--50\% to
$\pm$5--8\% after calibration. For visual embeddings, CLIP~ViT-B/32~\parencite{radford2021clip}
(345~MB, $\sim$900~ms) was replaced by MobileCLIP2-S2 (143~MB, $\sim$450~ms), a
2.4$\times$ smaller and 2$\times$ faster model with the same 512-D embedding
space and comparable semantic accuracy. A single YOLO11n-Seg model is reused for
both registration (background removal) and search (detection $+$ segmentation),
saving $\sim$19~MB versus a separate search detector and guaranteeing a shared
label space.

\subsection{INT8 quantization}
Depth Anything~V2 is a Vision Transformer dominated by \texttt{MatMul}/%
\texttt{Gemm} operators, an ideal candidate for dynamic INT8 weight
quantization~\parencite{jacob2018quantization}. As shown in Table~\ref{tab:quant}, quantization reduces model size
2.8$\times$, latency 2.4$\times$, and collapses the deployment to a single file
(embedding the external weights), at $\leq$2\% accuracy degradation. The
equivalent attempt on MobileCLIP2-S2 failed: its MobileOne backbone is dominated
by \texttt{Conv2D} operators, whose INT8 equivalent (\texttt{ConvInteger}) is not
implemented in ONNX Runtime on Android, yielding only a 15\% reduction; an FP16
export raised \texttt{Cast} node-type conflicts. MobileCLIP2-S2 therefore remains
FP32, documented as future work pending \texttt{ConvInteger} support.

\begin{table}[t]
\centering
\caption{INT8 quantization results for Depth Anything V2.}
\label{tab:quant}
\footnotesize
\begin{tabular}{@{}lccc@{}}
\toprule
\textbf{Criterion} & \textbf{FP32} & \textbf{INT8} & \textbf{Gain} \\
\midrule
Model size        & $\sim$100~MB $+$ data & 35~MB & 2.8$\times$ \\
Files needed      & 2 (model $+$ weights) & 1 (single file) & simplified \\
CPU latency       & $\sim$1200~ms         & $\sim$491~ms & 2.4$\times$ \\
Metric accuracy   & reference             & equivalent & $\leq$2\% drop \\
\bottomrule
\end{tabular}
\end{table}

\subsection{Custom banknote detector}
\begin{figure}[t]
\centering
\includegraphics[width=0.85\columnwidth]{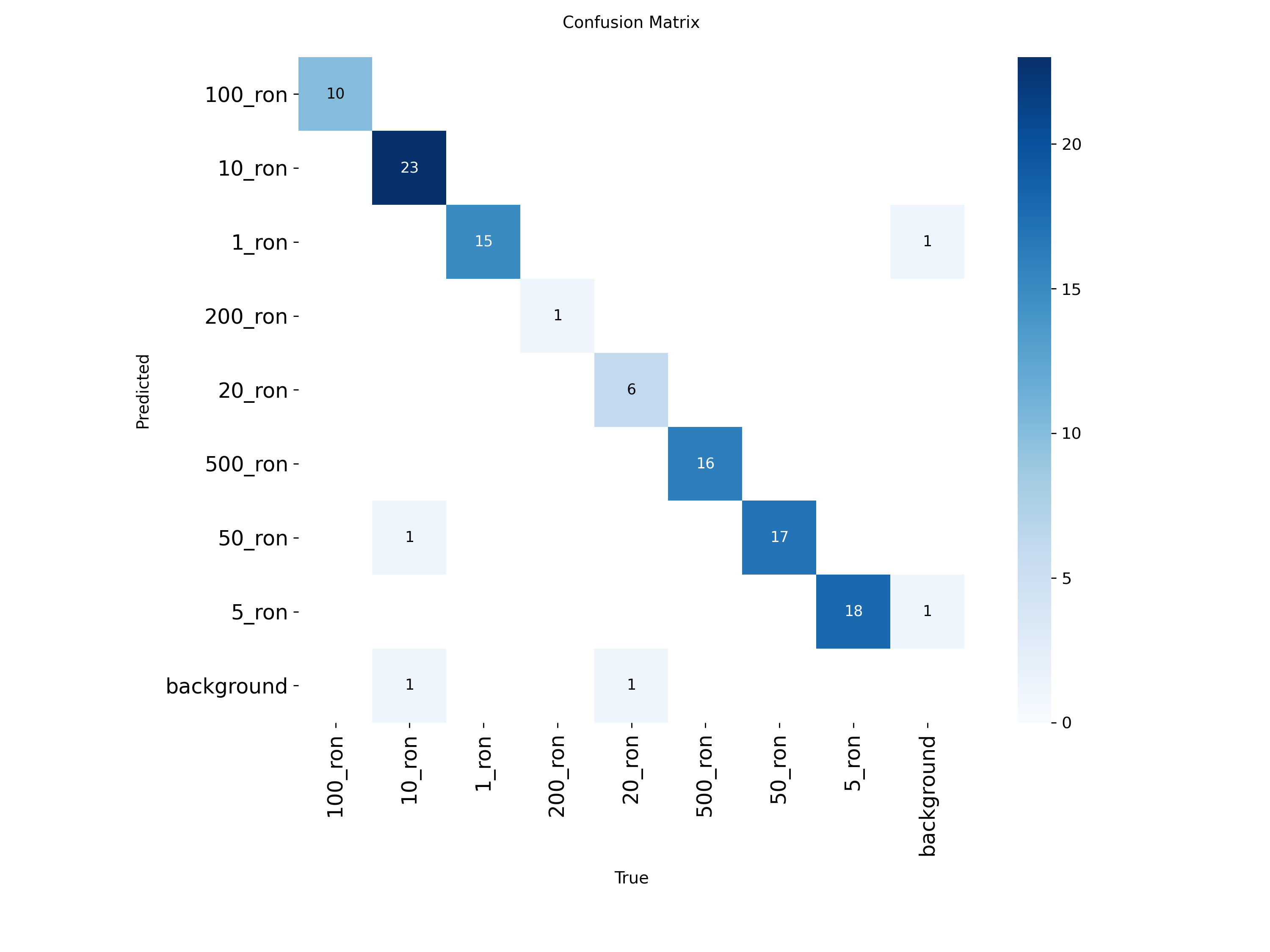}
\caption{Confusion matrix on the v3 validation set. The dominant diagonal
confirms high accuracy; 500~RON (absent from the v1 evaluation) is now
perfectly classified (16/16).}
\label{fig:money}
\end{figure}

Training stopped at epoch~85 via early stopping; the best epoch (60) reached a
mean average precision at IoU~0.50 (mAP@50) of \textbf{0.986} and mAP@50--95 of
\textbf{0.961}, with overall precision
0.975 and recall 0.925 on the validation set (Table~\ref{tab:money}). The
confusion matrix (Fig.~\ref{fig:money}) confirms that the v3 split rebalancing
fully resolved the 500~RON evaluation gap: all 16 validation instances are
classified correctly. The two residual error sources are an occasional
10~RON~$\leftrightarrow$~50~RON confusion (similar pink-violet palettes) and the
under-representation of 200~RON in validation (only two instances). Total
training took $\sim$38~min on a Tesla~T4; the exported ONNX model (9.8~MB) runs
in $<$15~ms per frame on the device.

\begin{table}[t]
\centering
\caption{Per-class detection metrics (YOLO26n v3, validation set).}
\label{tab:money}
\footnotesize
\begin{tabular}{@{}lcccl@{}}
\toprule
\textbf{Class} & \textbf{Inst.} & \textbf{Prec.} & \textbf{Recall} & \textbf{Note} \\
\midrule
\textbf{All}  & \textbf{106} & \textbf{0.975} & \textbf{0.925} & mAP@50 $=$ 0.986 \\
1 RON   & 15 & 0.938 & 1.000 & 1 FP from background \\
5 RON   & 18 & 0.947 & 1.000 & 1 FP from background \\
10 RON  & 25 & 0.958 & 0.920 & 1 confused w/ 50 RON \\
20 RON  & 6  & 1.000 & 1.000 & perfect \\
50 RON  & 17 & 0.944 & 1.000 & \\
100 RON & 10 & 1.000 & 1.000 & perfect \\
200 RON & 2  & 1.000 & 0.500 & under-represented \\
500 RON & 16 & 1.000 & 1.000 & resolved in v3 \\
\bottomrule
\end{tabular}
\end{table}

\subsection{On-device latency}
Table~\ref{tab:latency} reports per-module latency. The camera module uses frame
skipping (depth every third frame, cached in between) for a perceived 7--8~FPS;
AR search runs at $\sim$5~FPS, sufficient for a stationary object; the money
module, thanks to the ultra-compact YOLO26n, achieves the fastest response among
ML-based modules; color detection is effectively instantaneous.

\begin{table}[t]
\centering
\caption{Per-module latency on Samsung Galaxy S21 Ultra 5G.}
\label{tab:latency}
\footnotesize
\begin{tabularx}{\columnwidth}{@{}Llcc@{}}
\toprule
\textbf{Module} & \textbf{Main operation} & \textbf{Lat. (ms)} & \textbf{FPS} \\
\midrule
Camera        & Depth INT8 (every 3rd frame) & $\sim$491 & 7--8 \\
Objects: Add  & YOLO11n-Seg $+$ MobileCLIP    & $\sim$550 & $\sim$2 \\
Objects: Search & YOLO11n-Seg $+$ MobileCLIP  & $\sim$560 & $\sim$5 \\
People        & YuNet $+$ MobileFaceNet       & $<$80     & $\sim$12 \\
Money         & YOLO26n                        & $<$15     & $\sim$20 \\
Colors        & HSV (no ML)                    & $<$1      & $\sim$30 \\
\bottomrule
\end{tabularx}
\end{table}

\subsection{Depth calibration}
The metric model systematically over-estimates indoor distances by a constant
$\approx$1.82$\times$ factor, corrected by $f_c=0.55$. Calibrated against
ruler-measured ground truth at 0.5--3.0~m (Table~\ref{tab:calib}), the mean error
is below 1~cm and the maximum error 1~cm (at 1.0~m). At 1.5~m (the proximity-%
alert threshold), the calibrated model returns exactly 1.50~m, ensuring alerts
fire precisely at the configured distance.

\begin{table}[t]
\centering
\caption{Depth calibration results ($f_c = 0.55$).}
\label{tab:calib}
\footnotesize
\begin{tabular}{@{}ccccc@{}}
\toprule
\textbf{Real (m)} & \textbf{Raw (m)} & \textbf{Cal. (m)} & \textbf{Err (cm)} & \textbf{Err (\%)} \\
\midrule
0.50 & 0.91 & 0.50 & $+$0 & 0.0 \\
1.00 & 1.84 & 1.01 & $+$1 & 1.0 \\
1.50 & 2.73 & 1.50 & 0    & 0.0 \\
2.00 & 3.64 & 2.00 & 0    & 0.0 \\
2.50 & 4.52 & 2.49 & $-$1 & 0.4 \\
3.00 & 5.45 & 3.00 & 0    & 0.0 \\
\midrule
\textbf{Mean} &      &      & $<$1~cm & $<$0.3 \\
\bottomrule
\end{tabular}
\end{table}

\subsection{Accessibility}
The Android Romanian TTS engine reaches the first spoken word in under
200~ms, imperceptible for command feedback. A deduplication mechanism prevents
repeating an unchanged color or distance, and proximity alerts respect the
configured per-level debounce (300--800~ms). Voice navigation commands
(\emph{camera}, \emph{objects}, etc.) exceed 95\% recognition accuracy thanks to
their lexical specificity, while parameterized commands (\emph{search [object]})
reach $\sim$85\%, influenced by the variability of the dictated object name.

\section{Conclusion}
We presented \emph{VisionAId}, an Android application for users with visual
impairment that integrates five everyday functions (proximity alerting with metric
monocular depth and on-demand narrative scene description, registration and AR
retrieval of personal objects, color identification, recognition of known
people, and Romanian-banknote detection) behind an interface designed from the
outset for non-visual use, with Romanian voice commands, full speech synthesis,
and graduated haptics. The principal technical contribution is the on-device
integration of six deep-learning models that run simultaneously or sequentially
on commodity hardware, addressing concrete challenges in ONNX Runtime
lifecycle management, camera sharing between CameraX and ARCore sessions, INT8
quantization that cuts depth latency from $\sim$1200~ms to $\sim$491~ms, and
linear metric calibration with sub-centimeter mean error. The cloud
(Gemini Flash) is used only complementarily, preserving offline operation of
all critical functions.

Validation confirmed that total per-frame latency stays below 600~ms on the
reference device, with accuracy sufficient for the target use case: obstacle
detection below 1.5~m is precise within $\pm$1~cm, and the from-scratch YOLO26n
banknote detector reaches an mAP@50 of 0.986. Two limitations temper these
results: all measurements were obtained on a single high-end device, so
mid-range performance (where latencies are 2--3$\times$ higher) is only partially
characterized; and the system has not yet been evaluated in a formal user study
with participants who are blind or have low vision, which we regard as the ultimate
measure of its utility. The most promising future direction
is a persistent LLM-based conversational assistant maintaining a semantic memory
of the explored environment and supporting tool-calling over local detections. A
second priority is improving performance on mid-range devices, where latencies
are 2--3$\times$ higher: NNAPI-compatible export avoiding problematic operators,
distillation of the depth model, quantization-aware retraining of MobileCLIP, and
adaptive thermal/battery throttling. Finally, extending toward outdoor scenarios
and validating with real users from the community of people with visual impairment remain
important directions, as the ultimate measure is the confidence and autonomy the
end user perceives.

\printbibliography

\end{document}